%% file: main.tex
\documentclass{article}

\PassOptionsToPackage{numbers, sort&compress}{natbib}
\usepackage[preprint,eandd]{neurips_2026}
\usepackage[utf8]{inputenc}
\usepackage[T1]{fontenc}
\usepackage{hyperref}
\usepackage{url}
\usepackage{booktabs}
\usepackage{amsfonts}
\usepackage{nicefrac}
\usepackage{microtype}
\usepackage{xcolor}
\usepackage{graphicx}
\usepackage{algorithm}
\usepackage[noend]{algorithmic}
\usepackage{amsmath}
\usepackage{amssymb}
\usepackage{mathtools}
\usepackage{amsthm}
\usepackage{epsfig}
\usepackage{xspace}
\usepackage{multirow}
\usepackage{listings}
\usepackage{enumitem}
\usepackage[normalem]{ulem}

\lstdefinelanguage{JavaScript}{
  keywords={const, let, var, function, return, if, else, for, while, true, false, null, undefined, class, new, this},
  sensitive=true,
  comment=[l]{//},
  morecomment=[s]{/*}{*/},
  morestring=[b]',
  morestring=[b]"
}

\newcommand{\name}{GameGen-Verifier}

\theoremstyle{plain}

\theoremstyle{definition}

\theoremstyle{remark}

\newenvironment{denseitemize}{
\begin{itemize}[topsep=2.5pt, partopsep=0pt, leftmargin=1.5em]
  \setlength{\itemsep}{2.5pt}
  \setlength{\parskip}{0pt}
  \setlength{\parsep}{0pt}
}{\end{itemize}}

\begin{document}

\title{\name{}: Parallel Keypoint-Based Verification for LLM-Generated Games via Runtime State Injection}

\author{%
  {\normalfont Chaobo Jia$^{1\dagger}$, Ruipeng Wan$^{2\dagger}$, Ting Sun$^{3\dagger}$, Weihao Tan$^{4}$} \\
  {\normalfont Borui Wan$^{5}$, Yuxuan Tong$^{6}$, Guangming Sheng$^{5}$, Hong Xu$^{1}$} \\
  {\normalfont \small $^1$CUHK \quad $^2$HUST \quad $^3$Lionrock AI Lab \quad $^4$NTU \quad $^5$HKU \quad $^6$Independent Researcher} \\
  {\normalfont \small $^\dagger$Equal contribution}
}

\maketitle

\begin{abstract}
LLM-based game generation promises to turn natural-language specifications into executable games, but progress is limited by the lack of reliable automated verification.
Unlike conventional code generation, game correctness is defined over long-horizon interaction: a game may appear correct while violating core mechanics such as state updates, interaction rules, and phase transitions.
Existing Agent-as-a-Verifier approaches collapse verification into open-ended gameplay, making verdicts reachability-bound, time-consuming, coverage-limited, and sensitive to the agent's gameplay ability.

We present \name{}, an automated verification paradigm for LLM-generated games that decomposes a specification into \emph{verifiable keypoints} and grounds them into independent verification units.
Each unit patches the game runtime into a concrete target state, executes a bounded interaction, and judges the outcome against the keypoint assertion.
We implement \textsc{GGV-Harness}, a scalable agentic harness providing concurrency management, runtime isolation, and fault recovery.

On \textsc{VeriGame}, our dataset of 100 games across seven genres, \name{} achieves up to 92.2\% accuracy against human judgments versus 58.8\% for the coverage-enforced Agent-as-a-Verifier baseline, while reducing wall-clock time by up to $16.6\times$.
\end{abstract}

\input{tex/introduction}
\input{tex/related}
\input{tex/motivation}
\input{tex/design}
\input{tex/evaluation}

\vspace{-4pt}

\section{Limitations}
\label{sec:limitations}

\name{} currently operates on web-stack games (JavaScript, TypeScript, HTML) and has not been validated on Godot, Unity, or Unreal Engine, where LLM-based generation pipelines remain immature, making engine-native adapters a natural next step once such pipelines mature.
Keypoint extraction is also not guaranteed to be complete, and state construction requires white-box access to the implementation to identify its parameter structure, which limits applicability to games whose internal state resists programmatic inspection.
Finally, \name{} is falsification-oriented: a passing verdict means no executed keypoint exposed a violation, not that the game is formally proven correct.

\vspace{-4pt}

\section{Conclusion}
\label{sec:conclusion}

We presented \name{}, a verification paradigm for LLM-generated games via runtime state injection that decomposes a natural-language specification into independent, parallelizable verification units, each grounded in a concrete runtime state patch and checked through bounded interaction.
\textsc{GGV-Harness} makes this practical at scale through explicit concurrency control, runtime isolation, and fault recovery.
On \textsc{VeriGame}, our dataset of 100 LLM-generated games across seven genres, \name{} substantially improves agreement with expert human judgments while reducing wall-clock time relative to coverage-enforced Agent-as-a-Verifier baselines, and we view this approach as a practical step toward reliable game generation infrastructure.

\newpage

\def\UrlBreaks{\do\/\do-}
\bibliography{main}
\bibliographystyle{plain}

\newpage
\phantomsection
\section*{Broader Impact and Ethical Considerations}
\label{sec:impact}

\textbf{Broader impact.}
This work develops a verification paradigm for LLM-generated games, contributing to the infrastructure needed to make game generation reliable and measurable.
On the positive side, scalable verification lowers the barrier to game creation by enabling non-expert users to trust that generated games behave as intended, potentially democratizing access to game development.
On the negative side, more reliable verification could accelerate the automated generation of low-quality or derivative game content at scale.
We do not foresee significant risks beyond those already associated with LLM-based code generation in general.

\textbf{Ethical considerations.}
\textsc{VeriGame} is a dataset designed for benchmarking automated verification of LLM-generated games, not a redistribution of existing commercial games.
Its examples are released as anonymized natural-language specifications that describe gameplay mechanics and verification-relevant rules at an abstract functional level.
We do not redistribute proprietary game code, assets, or other expressive media from the source games; game identities are anonymized and references to original titles are removed before generation and evaluation.
These measures are intended to reduce copyright and trademark risks while preserving the research value of evaluating whether LLM-generated games satisfy functional specifications; our release should not be used to clone or market substitutes for existing games.
The dataset and verification framework are intended for research on reliable game generation, automated evaluation, and specification-grounded feedback, not to reproduce protected commercial content or evade the rights of game developers.
If concerns about a released specification are raised by rights holders, we will review the example and remove or revise it when appropriate.

\section*{Code and Dataset Licenses}

\textbf{Codebase.} The verification framework will be released under the MIT License upon acceptance.

\textbf{Datasets.} The \textsc{VeriGame} dataset will be released under CC BY 4.0 upon acceptance.

\newpage

\appendix
\input{tex/appendix}

\end{document}

%% file: tex/introduction.tex
\section{Introduction}

\begin{figure}[!t]
    \centering
    \includegraphics[width=\linewidth]{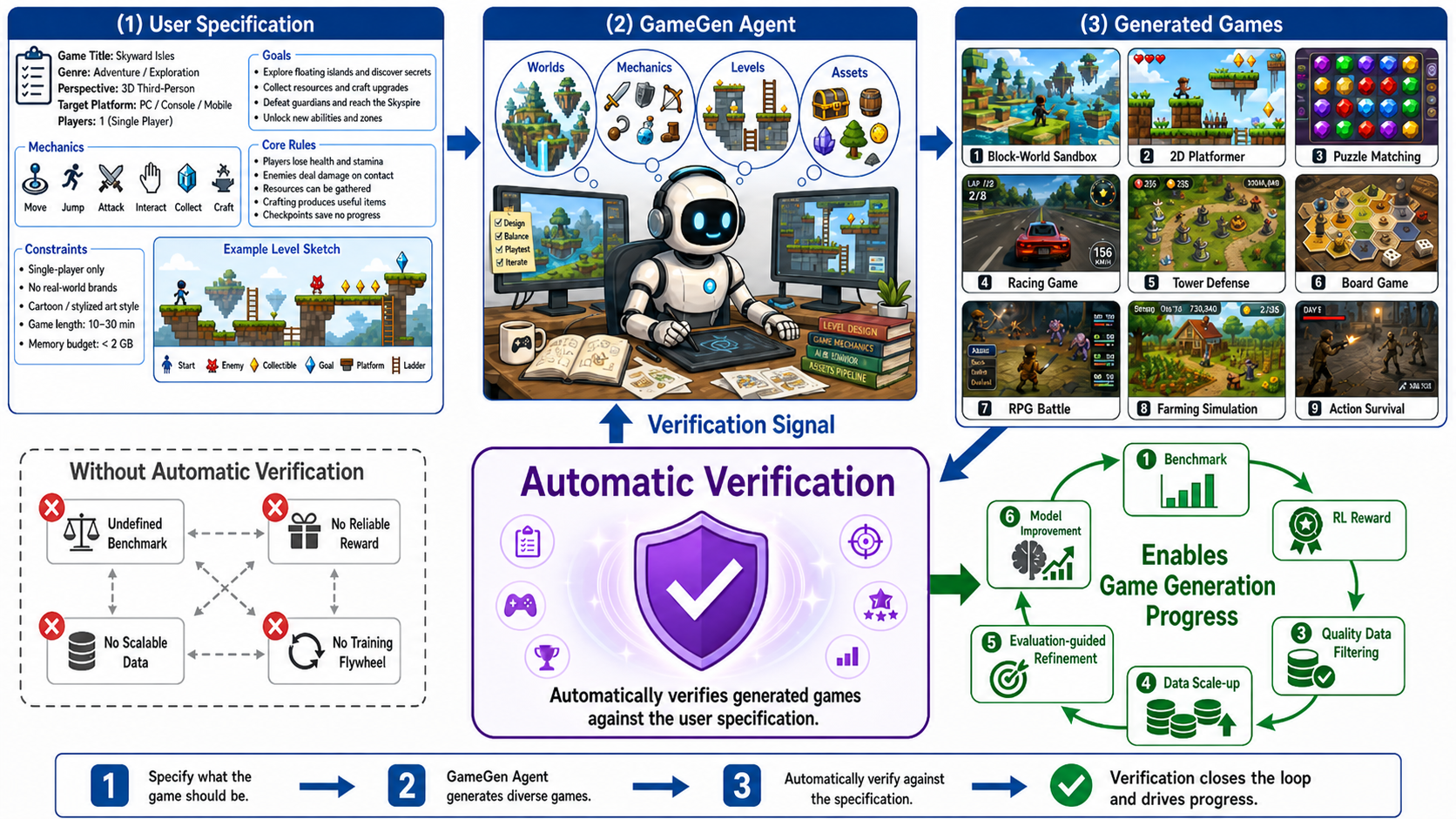}
    \vspace{-8pt}
    \caption{A user specification (1) drives a game generation agent that synthesizes worlds, mechanics, levels, and assets to produce playable games across diverse genres (2, 3). An automatic verifier then checks each generated game against the original specification and feeds correctness signals back to the agent, closing the generation loop and enabling downstream benchmarking, data curation, and model improvement (bottom right).}
    \vspace{-10pt}
    \label{fig:intro}
\end{figure}

\begin{figure}[!t]
\centering
\includegraphics[width=\linewidth]{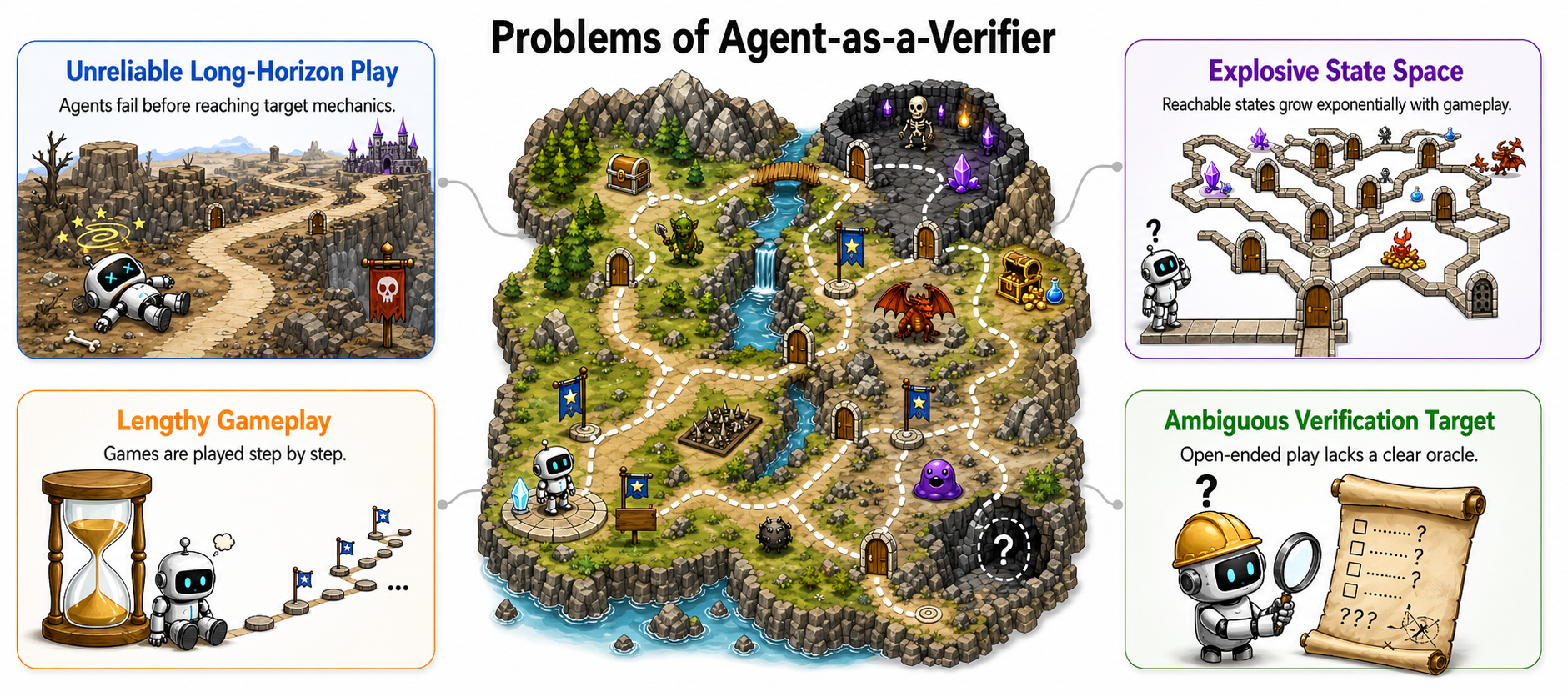}
\vspace{-12pt}
\caption{Applying AaaV to game verification forces the agent through long gameplay sequences to reach target mechanics, but the exponentially growing state space makes full coverage intractable and open-ended play provides no clear correctness oracle, making AaaV slow, coverage-limited, and misaligned with specification-level judgment.}
\vspace{10pt}
\label{fig:moti}
\end{figure}

Historically, turning a game proposal into a playable artifact required expert development teams, specialized tooling, and long production cycles.
Recent work has begun to explore LLM-based game generation systems that aim to rapidly turn natural-language specifications into executable, playable games~\cite{hong2024chatge,godogen,gallotta2024llmgames,anuttacon,opengame2026}.
However, progress in game generation is increasingly bottlenecked by the lack of reliable \emph{verification} of whether generated games correctly implement their specifications~\cite{opengame2026,godogen}. 
For conventional code generation, reliable tests enable benchmark construction, scalable data curation, and accurate reward signals that sustain the training flywheel~\cite{chen2021humaneval,hendrycks2021measuring,li2022competition,le2022coderl,liu2023rltf}.
For game generation, the analogous judge remains missing: a generated game may build successfully and appear visually correct while violating core mechanics such as state updates, interaction rules, and phase transitions.
Without a verifier that agrees with human judgment on runtime logical correctness, benchmark construction remains ill-defined, high-quality training data cannot be scaled, and training lacks trustworthy feedback.
Scalable verification is therefore a prerequisite for turning game generation from qualitative demonstrations into a well-defined research problem, as illustrated in Figure~\ref{fig:intro}. 
The closest existing paradigm is Agent-as-a-Verifier (AaaV), in which an autonomous agent interacts with a generated artifact at runtime and verifies the observed behavior against the specification~\cite{uxagent2024,webgenbench2025,mimo2026,minimax2025,webcompass2026}.
However, when applied directly to games, AaaV collapses verification into open-ended gameplay, with the limitations summarized in Figure~\ref{fig:moti}: gameplay rewards progress and exploration, whereas verification requires systematic coverage of specified mechanics and causal attribution of failures.
Before checking any target mechanic, the verifier must also reach a state where it is exercised, making evaluation time-consuming, coverage-limited, and dependent on agents that are themselves unreliable at complex gameplay.
Consequently, AaaV cannot serve as a scalable verifier for game generation.

To address this verification bottleneck, we make two key observations.
First, many specification-level failures in generated games can be exposed by a sparse set of localized critical behaviors, including build and launch health, collision handling, reward settlement, state updates, rule triggers, phase transitions, and progression changes, rather than by exploring entire gameplay trajectories.
Second, the states needed to exercise these behaviors need not be reached through gameplay: games are state machines whose runtime states are characterized by parameters, and for LLM-generated games, the verifier can identify this parameter structure from the white-box implementation and instantiate relevant states directly through runtime state-patching (Section~\ref{sec:construction})~\cite{chrome_devtools_runtime,godot_nodes_scene_instances,unity_object_instantiate,unreal_spawnactor}.

Building on these observations, we present \textbf{\name{}}, an automated verification paradigm for LLM-generated games.
At the core of \name{} is a keypoint abstraction: \name{} extracts sparse, specification-derived critical conditions from the specification and formulates them as \emph{verifiable keypoints}.
Each keypoint is a localized behavioral assertion, casting correctness as a local, bounded check rather than a global trajectory-level judgment.
To make these keypoints executable, \name{} grounds each one into a state-grounded verification unit: the verifier constructs a precise game state described by the implementation's parameter structure, injects it into the runtime, executes a bounded interaction sequence, and judges whether the observed outcome satisfies the expected one.
This formulation also makes verification units self-contained, reducing unreliable gameplay to a finite set of parallelizable short-horizon verifications.

\name{} closes the loop by attributing keypoint verdicts back to specification elements and propagating \textsc{fail} verdicts through their dependency structure.

We further propose \textsc{GGV-Harness}, a scalable agentic verification harness built on a two-layer orchestration-worker architecture that enforces concurrency management, runtime isolation, and fault recovery.

This paper makes the following contributions:
\begin{denseitemize}
    \item We identify the verification bottleneck in game generation and show why existing paradigms are insufficient for scalable correctness judgments.
    \item We propose \name{}, a new verification paradigm that reformulates game verification from open-ended trajectory exploration into state-grounded verification of specification-derived verifiable keypoints.
    \item We implement a scalable agentic verification harness centered on three systems guarantees: concurrency management, runtime isolation, and fault recovery, enabling reliable large-scale verification execution.
    \item We validate \name{} on \textsc{VeriGame} and show that it achieves up to 92.2\% Acc@5 and 95.4\% F1@5 against human judgment, while reducing wall-clock time by up to $16.6\times$ compared with the coverage-enforced Agent-as-a-Verifier baseline.
\end{denseitemize}

%% file: tex/related.tex
\vspace{-8pt}

\section{Related Work}

\paragraph*{Evaluation paradigms and benchmarks for code generation.}
Evaluation of code generation has evolved alongside task complexity.
Early benchmarks relied on deterministic test-based verification, where correctness is decided by predefined input-output test cases~\cite{chen2021humaneval,austin2021program,hendrycks2021measuring,jimenez2023swebench}.
As tasks expanded to full-stack systems, static tests became insufficient, leading to LLM-as-a-Judge approaches that score generated artifacts against human-defined rubrics~\cite{llmasjudge2023,li2024prompting,webgenbench2025}.
More recently, Agent-as-a-Verifier (AaaV) approaches have emerged for interactive systems such as web applications, where autonomous agents interact with the live system to verify functionality~\cite{uxagent2024,webgenbench2025,mimo2026,minimax2025,webcompass2026}.
AaaV is the closest existing evaluation paradigm to our setting because it verifies runtime behavior rather than static code alone.
However, existing AaaV benchmarks are primarily designed for web applications, where states are largely discrete, interactions are short, and verification targets are typically reachable through a small number of UI actions.

\paragraph*{Game testing.}
Game testing has a long tradition in both industrial practice and software-engineering research.
Surveys of video-game testing report that production pipelines still rely heavily on manual playtesting and testers' domain knowledge, while automated techniques remain difficult to generalize across genres and development workflows~\cite{politowski2021survey}.
Representative approaches include runtime or model-based monitoring, which instruments the game loop or an abstract model to check temporal gameplay properties and expose bugs~\cite{varvaressos2017automated}, and agent-based automated playtesting, where reinforcement-learning or MCTS-style agents generate play traces to search for defects or coverage gaps~\cite{ariyurek2021automated,artificial_playfulness}.
In the context of LLM-generated games, Godogen~\cite{godogen} incorporates game-specific tests such as task-specific harnesses, execution-time assertions, and screenshot-based visual QA during the development process, while OpenGame~\cite{opengame2026} evaluates agentic web-game generation through build, visual usability, and intent-alignment signals.
These techniques are useful for validating existing games or local development-time properties, but they typically assume a human-authored target game, a handcrafted model or oracle, or a tester agent that still has to traverse gameplay states.
They do not directly address scalable specification-level fidelity checking for games generated from natural language.

\paragraph*{Coding agents.}
LLM-powered coding agents~\cite{codex,claude_code,open_code} have moved programming from one-shot synthesis to iterative agent-driven development within real execution environments, supporting workflows in which models inspect repositories, edit files, execute commands, and refine outputs based on environmental feedback.
Multi-agent and agent-swarm systems further decompose tasks into sub-problems for coordinated parallel execution~\cite{autogen,metagpt,tot,kimik25}.
Our agentic verification workflow draws on this paradigm but is scenario-specific and vendor-independent, designed for stable parallel evaluation rather than general-purpose task orchestration.

%% file: tex/motivation.tex
\vspace{-8pt}

\section{Background}
\label{sec:background}

\paragraph*{LLM-based Game Generation.}
LLM-based game generation aims to turn a natural-language game proposal into an executable, playable artifact~\cite{hong2024chatge,godogen,gallotta2024llmgames,anuttacon,opengame2026}.
In our setting, the input is a \emph{specification}: a natural-language document that describes the intended game, covering its objective, entities, player controls, rules, scoring or reward logic, failure conditions, progression structure, and other aspects.
The generator is then asked to realize this specification as a complete interactive system, synthesizing code, assets, and supporting structures for runtime loops, entity and state management, gameplay logic, input handling, collision or physics behavior, UI, and other game-specific components.

\paragraph*{Game-generation Stack.}
Recent LLM-based game generation systems largely target web-stack games built in JavaScript, TypeScript, and HTML~\cite{godogen,opengame2026}. 
Mainstream engines such as Godot, Unity, and Unreal Engine are beginning to receive community MCP (Model Context Protocol) integrations that expose scene and object inspection, node or object creation, and script editing to agents~\cite{godot_mcp_ee0pdt,unity_mcp_coplaydev,unreal_mcp_chir24}, but this support remains fragmented and immature, making these engines difficult to use as routine generation stacks.

\paragraph*{Hoare triples.}
Hoare logic is a classical formalism for reasoning about program correctness through triples of the form $\{P\}\,C\,\{Q\}$~\cite{hoare1969}.
Here, $P$ is a precondition over the program state, $C$ is a command or program fragment to be executed, and $Q$ is a postcondition over the resulting state.
Under the standard partial-correctness interpretation, the triple asserts that if execution starts in any state satisfying $P$ and $C$ terminates, then the resulting state satisfies $Q$.
Hoare triples therefore provide a compact way to specify program behavior by relating assumptions before execution to guarantees after execution.
In our gamegen-verification setting, we use this structure in a Hoare-style sense: the command component $C$ is specialized to a bounded game-interaction sequence $\mathbf{a}$.

%% file: tex/design.tex
\vspace{-8pt}
\section{\name{}}

\begin{figure}[t]
\centering
\includegraphics[width=\linewidth]{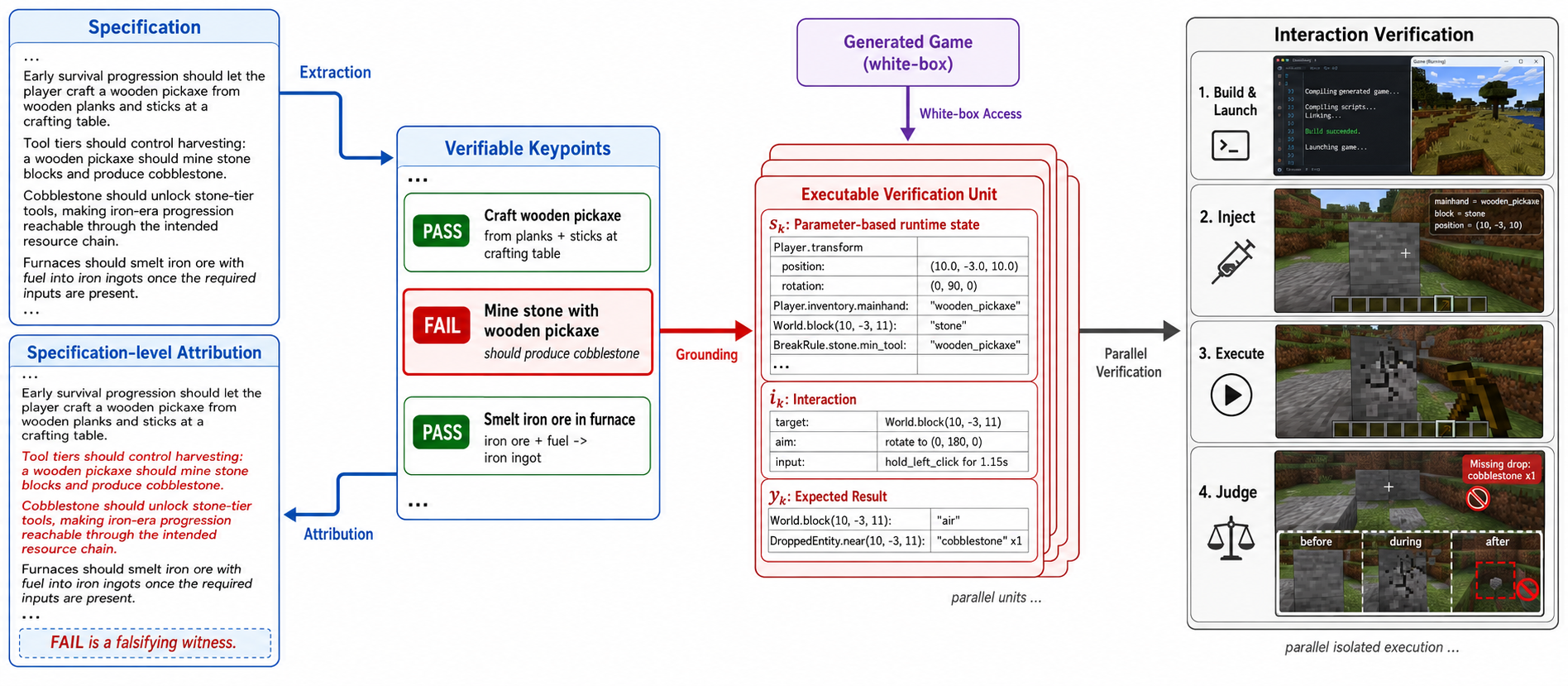}
\vspace{-18pt}
\caption{\name{} extracts verifiable keypoints from a natural-language specification and grounds each into an executable verification unit comprising a parameter-based runtime state $s_k$, a bounded interaction $i_k$, and an expected outcome $y_k$, constructed via white-box access to the generated game. Each unit is executed through four steps: build and launch, inject the target state, execute the bounded interaction, and judge the observed outcome. \textsc{Fail} verdicts propagate back to the specification as falsifying witnesses.}
\vspace{-6pt}
\label{fig:design}
\end{figure}

\name{} introduces a state-grounded verification workflow (Figure~\ref{fig:design}, Section~\ref{sec:keypoint}), proposes a parameter-based state construction and runtime injection mechanism to make keypoint verification executable (Section~\ref{sec:construction}), and implements \textsc{GGV-Harness}, a scalable agentic verification harness for concurrency management, runtime isolation, and fault recovery (Section~\ref{sec:workflow}).

\vspace{-8pt}

\subsection{Verifiable Keypoint Extraction and Attribution}
\label{sec:keypoint}

In LLM-generated games, the natural-language specification defines what the game should be.
We observe that many specification-level failures can be exposed by a sparse set of critical behavioral conditions, often located at cross-system interactions such as collision and physics responses, inventory or resource updates, scoring and reward settlement, and progression gates.
We therefore extract these conditions as verifiable keypoints and test them against the generated implementation, using their verdicts to derive specification-level judgments without traversing the full state space.

To make this notion precise, we define a \emph{verifiable keypoint} as a natural-language Hoare-style triple $\varphi = (P, \mathbf{a}, Q)$, where $P$ is a precondition over game states, $\mathbf{a}$ is a bounded interaction sequence, and $Q$ is the expected postcondition.
Each keypoint must satisfy three conditions:
\textbf{(C1)~Constructibility}: $P$ describes a directly constructible game state;
\textbf{(C2)~Boundedness}: $\mathbf{a}$ involves a finite, short sequence of actions;
\textbf{(C3)~Verifiability}: $Q$ is unambiguous enough to admit an operational \textsc{pass}/\textsc{fail} verdict.
We use an LLM agent to extract candidate verifiable keypoints from the specification under the C1--C3 constraints, representing each accepted keypoint as a triple $(P,\mathbf{a},Q)$.
This yields a finite set of verifiable keypoints from each game specification, framing correctness as a local, bounded property rather than a global trajectory property.

A verifiable keypoint as defined above is still a natural-language assertion: it states what should hold, but is not directly runnable.
To make it executable against a concrete game implementation, we introduce a \emph{verification unit} $u_k = (s_k, i_k, y_k)$ as the concrete runtime grounding of a keypoint.
Compared with the keypoint triple $(P,\mathbf{a},Q)$, the verification unit grounds each natural-language component into an implementation-specific counterpart.
The state $s_k$ instantiates the precondition $P$ via our parameter-based state construction mechanism (Section~\ref{sec:construction}).
Given $s_k$, the interaction $\mathbf{a}$ is grounded into an executable instruction $i_k$, and the postcondition $Q$ is operationalized into a concrete expected outcome $y_k$.
A single verifiable keypoint may be grounded into multiple verification units to cover distinct boundary conditions (e.g., testing collision under varying positions and velocities).
Each unit is an independent, bounded evaluation: the verifier launches an isolated game instance, injects constructed state $s_k$ into the runtime, executes $i_k$, and collects runtime evidence, including visual traces (e.g., screenshots) and internal runtime signals (e.g., logs or exposed state).
The bounded evidence is judged against $y_k$ by a vision-language model (VLM) judge, supplemented by programmatic assertions when the expected outcome can be expressed as predicates over exposed runtime state or logs.
A unit receives a \textsc{fail} verdict if the generated game fails to build or launch, fails to realize the injected state after the harness's retry procedure, fails to complete the bounded interaction, or produces behavior inconsistent with the expected outcome.
Each verification unit is self-contained, eliminating inter-unit dependencies and naturally enabling parallel execution.
This state-grounded formulation avoids the reachability problem of ordinary gameplay, reducing unreliable gameplay to a finite collection of parallelizable short-horizon verifications.

The specification is the scoring interface for both human and automated verification, so \name{} closes the loop from specification to keypoints and back: each keypoint verdict is attributed to the specification element it covers and aggregated into a specification-level judgment.
This specification-level aggregation follows a falsification-oriented scoring rule.
A failed keypoint is treated as a falsifying witness for the specification element it covers, and the failure is propagated to downstream elements whose satisfaction depends on that behavior.
After failure propagation, the remaining specification elements are assigned \textsc{pass}.
This means that no executed keypoint has exposed a violation of those elements, not that the generated game is formally proven correct on them.

\name{} accordingly exports results at two levels.
\emph{Keypoint-level results} record each verifiable keypoint's \textsc{pass}/\textsc{fail} verdict evaluated independently via constructed states.
\emph{Specification-level results} aggregate keypoint verdicts back to specification elements.

\vspace{-8pt}

\subsection{Parameter-Based State Construction and Runtime Injection}
\label{sec:construction}
To ground a verifiable keypoint into an executable verification unit $u_k = (s_k, i_k, y_k)$, the verifier must instantiate a target state $s_k$ satisfying the keypoint precondition $P$.
For games, this is challenging because states are large, continuous, path-dependent, and often reachable only after long interaction histories.
If the verifier must reach $s_k$ through gameplay, verification again becomes the long-horizon exploration problem that \name{} is designed to avoid.
We therefore reframe this prerequisite from a \emph{reachability problem} (whether there exists an interaction trajectory from $s_0$ to a state satisfying $P$) to a \emph{configuration problem}: directly construct a target state satisfying $P$ from the game's parameter structure.

Our key insight is that \emph{LLM-generated games offer an asymmetry unavailable in traditional game evaluation}: the verifier has access not only to the deployed game, but also to the full implementation source produced by the generator.
While the deployed game exposes only an interactive surface, the source code reveals the parameters governing its internal state.
Since interactive software can be viewed as a data-driven state machine, a target game state can be characterized by setting the relevant entities, variables, flags, and relations.
\name{} exploits this structure by synthesizing state snapshots that satisfy each target precondition and injecting them into the runtime, localizing verification to a controlled starting point.

For example, instead of requiring a verifier to discover a boss fight through exploration, \name{} can directly construct a state in which the boss is spawned, the player is placed in the correct arena, quest flags are advanced, and combat parameters are set to the desired phase.
This converts dynamic exploration into static state construction.
This distinction is especially important for games and other long-horizon interactive software, whereas many web tasks expose short, discrete, and easily reachable states that do not require parameter-based construction.

To put this construction into practice, the verifier must instantiate synthesized states directly in the runtime environment.
Existing checkpointing or save/load mechanisms are often engine-specific, incomplete, or otherwise unsuitable for systematic evaluation.
We therefore formalize an engine-agnostic injection contract: the verification harness requires access to a \emph{runtime state patching} mechanism, i.e., programmatic modification of the game's internal state after launch without replaying interaction history.
This requires two basic capabilities: controlling \emph{what entities exist} in the running scene (e.g., loading a level, spawning or removing objects) and controlling \emph{what values they hold} (e.g., setting health, advancing quest flags, or repositioning characters).
The constructed state need not reproduce transient execution context such as in-flight coroutines or animation frames exactly.
It only needs to be \emph{logically equivalent} with respect to the keypoint precondition and expected outcome, since verification postconditions are defined over observable behavior.

Fortunately, web-stack games often expose mutable JavaScript state through the browser runtime or injected instrumentation, making runtime state patching comparatively straightforward~\cite{chrome_devtools_runtime}.
Our experiments therefore focus on web-stack games.
As discussed in Section~\ref{sec:background}, mainstream engines such as Godot, Unity, and Unreal Engine are outside our experimental scope because LLM-based generation for these engines remains an open problem.
We discuss them as contract-compatible extensions, since they also support runtime state patching mechanisms~\cite{godot_nodes_scene_instances,unity_object_instantiate,unreal_spawnactor}.
We discuss engine-specific mappings in Appendix~\ref{app:engine}.

\vspace{-8pt}

\subsection{Scalable Agentic Verification Harness}
\label{sec:workflow}
A natural implementation of \name{} is to run the entire workflow inside a modern coding-agent session, such as Claude Code or Codex.
In this design, a parent agent is equipped with multiple task-specific skills, dispatches verification units to built-in subagents, and later merges the returned results.
However, this agent-native harness is fragile as the execution substrate for the full workflow.
\name{} must evaluate many units across games, keypoints, and repeated trials, while current sub-agent mechanisms provide limited support for parallel scheduling, explicit concurrency control, fault-tolerant recovery, and runtime isolation.
The core systems challenge is therefore reliable execution management for a highly parallel verification workload.

We implement \textsc{GGV-Harness} as a two-layer agentic verification harness that separates orchestration from local reasoning.
The Python \emph{orchestration layer} maintains the unit queue, enforces concurrency and rate limits, applies timeout and retry policies, persists checkpoints, and aggregates verdicts.
The \emph{worker layer} consists of short-lived LLM agent calls, each scoped to one self-contained verification unit $u_k=(s_k,i_k,y_k)$, such as constructing an injectable state, generating an interaction script, or judging observed behavior from runtime evidence.
Because units are independent, workers can be invoked statelessly: each processes one verification unit and returns structured evidence and a verdict.
The orchestrator can then dispatch units concurrently, retry failed or timed-out units, and resume interrupted runs from checkpoints.
To provide runtime isolation, each worker launches an independent game instance, preventing interference through shared state such as random seeds, timers, or browser-level singletons.
Together, this design retains coding agents for local reasoning while providing the concurrency control, runtime isolation, and fault recovery needed for large-scale, repeatable verification.

%% file: tex/evaluation.tex
\section{Evaluation}
We validate \name{} through two sets of experiments: a verification study against human judgment and a harness study that isolates the scalability and stability benefits of \textsc{GGV-Harness}.

\subsection{Dataset}
We build \textsc{VeriGame}, a game specification dataset for automated verification of LLM-generated games, since no such dataset is currently available.
\textsc{VeriGame} contains 100 game-generation tasks spanning seven genres (Action, Adventure, Casual, Puzzle, Simulation, Strategy, and Board), each consisting of an anonymized game identifier and a natural-language specification.
The dataset is constructed through a strict sourcing, specification synthesis, and human refinement pipeline.
We provide full details of the data synthesis pipeline in Appendix~\ref{sec:data}. 

\subsection{Experimental Setup}
\textbf{Game-generation Stack.}
As discussed in Section~\ref{sec:background}, current LLM-based game generation for mainstream engines such as Godot, Unity, and Unreal Engine remains immature.
These engines are therefore outside the experimental scope of this work.
We therefore evaluate web-stack games implemented in JavaScript, TypeScript, or HTML, using each game's specification from \textsc{VeriGame} as input.
Games are generated by the coding agent: Claude Code v2.1.72 (Opus 4.6, high effort).

\textbf{Baselines.}
We compare against two Agent-as-a-Verifier (AaaV) variants~\cite{uxagent2024,webgenbench2025,mimo2026,minimax2025,webcompass2026}, the current state-of-the-art paradigm for evaluating LLM-generated interactive software, which has been widely adopted for web application verification.
\emph{AaaV-Direct} is the standard prompt-only baseline: an evaluation agent is prompted to directly play the generated game and verify the observed behavior against the full natural-language specification.
This setting reflects the common use of AaaV, but it may omit hard-to-reach or low-salience specification elements.
We therefore also evaluate \emph{AaaV-CE} (coverage-enforced AaaV), which wraps the same direct verifier in a todo-list-style coverage harness over the specification elements.
After each agent response, a judge checks whether every specification element has an explicit verdict.
If any element is missing or ambiguous, the harness forces the agent to continue until the list is complete or the retry budget is exhausted.
This makes AaaV-CE a stronger AaaV baseline than AaaV-Direct.
The full AaaV-Direct prompt is provided in Appendix~\ref{app:AaaV_prompt}.
All methods produce specification-level verdicts, enabling direct comparison.

\subsection{Alignment with Human Judgment}
\label{sec:alignment}
This experiment evaluates whether \name{} produces verdicts that agree with human judgment and whether it offers efficiency gains over AaaV-style verification once explicit specification-level coverage is required.
We evaluate \name{}, AaaV-Direct, and AaaV-CE under three underlying-agent settings: Codex CLI v0.125.0 (GPT-5.4 xhigh), Claude Code v2.1.72 (Opus 4.6, high effort), and Claude Code v2.1.72 (Kimi 2.6 Coding).
In \name{}, the agent serves as the \textsc{GGV-Harness} worker.
AaaV-Direct uses the same agent to directly play the generated game and executes verification against the full specification.
AaaV-CE places the same direct-verification agent inside the todo-list coverage harness, which repeatedly judges whether all specification elements have verdicts and forces continuation when the list is incomplete.
For each specification in \textsc{VeriGame}, we generate a web-stack implementation and run each automated method $5$ times for each generated game under each setting.
As discussed in Section~\ref{sec:keypoint}, the specification serves as a unified scoring interface: all automated methods are ultimately scored on the same specification elements.
The human reference and \name{} produce binary specification-level labels, \textsc{pass} or \textsc{fail}.
The AaaV baselines may additionally return \textsc{unverified} when a specification element is not verified or the agent provides no supporting evidence for a binary verdict.

\paragraph*{Human Reference.}
As the human reference, $3$ expert human evaluators independently play each generated game and assess its correctness against the specification.
Each evaluator plays the game and marks which specification elements the implementation fails to satisfy based on observed gameplay behavior.
The final human verdict is determined by majority vote across the evaluators.
This produces the same specification-level verdicts as these automated methods.

\paragraph*{Metrics.}
All automatic metrics are computed over $5$ repeated runs, using \textsc{pass} as the positive class.
We report five metrics against the human reference:
\emph{Acc@5} measures overall label agreement, \emph{Prec@5} measures the correctness of predicted \textsc{pass} labels, \emph{Rec@5} measures coverage of human-\textsc{pass} elements, \emph{F1@5} is the harmonic mean of precision and recall, and \emph{Time@5} measures wall-clock execution time.
Appendix~\ref{app:metrics} gives the full formulas, including how AaaV \textsc{unverified} outcomes are scored.

\begin{table*}[t]
\centering
\setlength{\belowcaptionskip}{3pt}
\caption{Results of alignment with human judgment across different verification methods and agent settings. Prec@5 penalizes false \textsc{pass} predictions, Rec@5 penalizes missed human-\textsc{pass} elements, and Time@5 reports average wall-clock time over five runs.}
\label{tab:e2e_alignment}
\small
\begin{tabular*}{\textwidth}{@{\extracolsep{\fill}}llccccc@{}}
\toprule
Backend & Method & Acc@5 & Prec@5 & Rec@5 & F1@5 & Time@5 \\
\midrule
Codex & AaaV-Direct & 0.011 & 0.667 & 0.009 & 0.018 & \textbf{387s} \\
Codex & AaaV-CE     & 0.588 & \textbf{0.968} & 0.484 & 0.645 & 7356s \\
Codex & \name{}     & \textbf{0.922} & 0.912 & \textbf{1.000} & \textbf{0.954} & 443s \\
\midrule
Claude Code (Opus) & AaaV-Direct & 0.013 & 0.714 & 0.011 & 0.022 & 552s \\
Claude Code (Opus) & AaaV-CE     & 0.576 & \textbf{0.935} & 0.475 & 0.630 & 8245s \\
Claude Code (Opus) & \name{}     & \textbf{0.902} & 0.892 & \textbf{1.000} & \textbf{0.943} & \textbf{513s} \\
\midrule
Claude Code (Kimi) & AaaV-Direct & 0.033 & 0.812 & 0.021 & 0.040 & 537s \\
Claude Code (Kimi) & AaaV-CE     & 0.510 & \textbf{0.923} & 0.522 & 0.667 & 9072s \\
Claude Code (Kimi) & \name{}     & \textbf{0.866} & 0.893 & \textbf{0.962} & \textbf{0.926} & \textbf{614s} \\
\bottomrule
\end{tabular*}
\vspace{2pt}
\raggedright
\end{table*}

\paragraph*{Results.}

Table~\ref{tab:e2e_alignment} shows that \name{} is more accurate and faster than AaaV-style verification.
\name{} achieves Acc@5/F1@5 of $0.922/0.954$, $0.902/0.943$, and $0.866/0.926$ under the Codex, Claude Code (Opus), and Claude Code (Kimi) settings, respectively.
AaaV-Direct has very low accuracy because it often terminates after checking only a small, easy subset of the specification, while AaaV-CE reaches at most $0.588$ Acc@5, far below \name{}.
Our trajectory inspection shows the reason: it can often verify directly observable, short-horizon specification elements, but it still struggles with elements that require sustained gameplay, precise state reachability, or multi-step causal attribution.

Across Codex, Claude Code (Opus), and Claude Code (Kimi), \name{} consistently outperforms AaaV-CE; Codex and Claude Code (Opus) are closely matched, and Claude Code (Kimi) remains substantially above its AaaV-CE counterpart.
The precision/recall pattern, especially \name{}'s high recall ($1.000$ under Codex and Claude Code (Opus), $0.962$ under Claude Code (Kimi)), also reflects the falsification-oriented design of \name{}: keypoint extraction can lose coverage, but observed keypoint violations provide strong evidence of real specification failures.
Finally, \name{} substantially reduces wall-clock time by up to $16.6\times$ compared with AaaV-CE by converting verification into local, bounded units that can be executed in parallel.

\subsection{Agentic Verification Harness}
\label{sec:harness_eval}
 
This experiment evaluates the scalability and execution guarantees of \textsc{GGV-Harness} as an execution harness for state-grounded verification.

We compare \textsc{GGV-Harness} with \emph{Agent-Native Harness}, a strong agent-native execution baseline described in Section~\ref{sec:workflow}.
Agent-Native Harness implements the same state-grounded verification workflow entirely inside a modern coding-agent session.
The parent agent is equipped with task-specific verification skills for keypoint grounding, state construction and injection, interaction-script generation, runtime evidence collection, and verdict aggregation.
To exploit unit independence, the parent agent uses its built-in subagent mechanism to dispatch multiple verification units in parallel, waits for the returned evidence and verdicts, and merges them into a game-level report.
To isolate the effect of the harness design, we run both harnesses under two settings: Codex CLI v0.125.0 (GPT-5.4 xhigh) and Claude Code v2.1.72 (Opus 4.6, high effort), and keep the verification workload and evaluation budget identical across methods within each setting: the same generated games, extracted keypoints, retry budget, and timeout policy are used for both methods.
We then compare how each harness controls concurrency, isolates runtime executions, and recovers from failures.

\paragraph*{Metrics.}
We report wall-clock interaction time for completed verification runs and separately record whether each execution mode provides the core systems guarantees required by Section~\ref{sec:workflow}: concurrency management, runtime isolation, and fault recovery.
Both methods are evaluated under the same timeout budget.

\begin{table*}[t]
\centering
\setlength{\belowcaptionskip}{3pt}
\caption{Harness-level scalability and execution-guarantee comparison. \textsc{GGV-Harness} is the only execution substrate that combines low wall-clock time with explicit concurrency control, runtime isolation, and fault recovery.}
\label{tab:harness_eval}
\small
\setlength{\tabcolsep}{3pt}
\begin{tabular*}{\textwidth}{@{\extracolsep{\fill}}llcccc@{}}
\toprule
Method & Backend & Time & Concurrency Control & Runtime Isolation & Fault Recovery \\
\midrule
Agent-Native Harness & Codex & 1293s & Partial & $\times$ & $\times$ \\
\textsc{GGV-Harness} & Codex & \textbf{457s} & \checkmark & \checkmark & \checkmark \\
\midrule
Agent-Native Harness & Claude Code & 1594s & Partial & $\times$ & $\times$ \\
\textsc{GGV-Harness} & Claude Code & \textbf{536s} & \checkmark & \checkmark & \checkmark \\
\bottomrule
\end{tabular*}
\vspace{2pt}
\raggedright
\end{table*}

\paragraph*{Results.}
Table~\ref{tab:harness_eval} shows that scalable verification requires more than invoking more agent workers.
Agent-Native Harness can invoke subagents, but execution is still coordinated through the parent agent session rather than through an explicit scheduler with controlled resource limits.
As a result, runtime isolation and recovery from partial failures are best-effort behaviors rather than execution guarantees.
By contrast, \textsc{GGV-Harness} reduces average wall-clock time by up to $66.4\%$ relative to Agent-Native Harness while adding the guarantees needed for repeatable evaluation.
Each worker agent is short-lived and bound to one verification unit, so failed or timed-out units can be retried without invalidating the rest of the game-level run.
Per-unit runtime isolation also prevents interference through shared browser state, timers, random seeds, or mutated game globals.

%% file: tex/appendix.tex
\section*{Appendix}

\section{Metric Definitions}
\label{app:metrics}

All alignment metrics are computed at the specification-element level against a binary human reference label $r \in \{\text{\textsc{pass}}, \text{\textsc{fail}}\}$.
\name{} also returns binary specification-level labels.
The AaaV baselines may additionally return $\text{\textsc{unverified}}$ when a specification element is not checked or the evidence is insufficient.
We treat $\text{\textsc{pass}}$ as the positive class.

We define the extended confusion counts as
\[
\begin{array}{c|cc}
 & r=\text{\textsc{pass}} & r=\text{\textsc{fail}} \\
\hline
\hat{y}=\text{\textsc{pass}} & TP & FP \\
\hat{y}=\text{\textsc{fail}} & FN & TN \\
\hat{y}=\text{\textsc{unverified}} & \mathrm{Unverified}_{+} & \mathrm{Unverified}_{-} \\
\end{array}
\]
and let $n = TP + FP + FN + TN + \mathrm{Unverified}_{+} + \mathrm{Unverified}_{-}$.
For binary-output methods such as \name{}, $\mathrm{Unverified}_{+}=\mathrm{Unverified}_{-}=0$, and the metrics reduce to the standard binary definitions:
\[
\mathrm{Acc} = \frac{TP + TN}{TP + FP + FN + TN},
\quad
\mathrm{Prec} = \frac{TP}{TP + FP},
\quad
\mathrm{Rec} = \frac{TP}{TP + FN},
\]
\[
\mathrm{F1} = \frac{2 \cdot \mathrm{Prec} \cdot \mathrm{Rec}}{\mathrm{Prec} + \mathrm{Rec}}.
\]

For AaaV-Direct and AaaV-CE, \textsc{unverified} outcomes are retained rather than removed from scoring.
They count as non-agreements for accuracy, and \textsc{unverified} outcomes on human-\textsc{pass} elements count as missed positives for recall:
\[
\mathrm{Acc} = \frac{TP + TN}{n},
\quad
\mathrm{Prec} = \frac{TP}{TP + FP},
\quad
\mathrm{Rec} = \frac{TP}{TP + FN + \mathrm{Unverified}_{+}},
\]
\[
\mathrm{F1} = \frac{2 \cdot \mathrm{Prec} \cdot \mathrm{Rec}}{\mathrm{Prec} + \mathrm{Rec}}.
\]

Time@5 is the wall-clock execution time averaged over the five repeated runs.

\section{Engine-Specific Runtime State Patching}
\label{app:engine}

As discussed in Section~\ref{sec:construction}, our verification paradigm requires the generated implementation to support runtime state patching: the ability to programmatically control what entities exist in a running scene and what values they hold~\cite{chrome_devtools_runtime,godot_nodes_scene_instances,unity_object_instantiate,unreal_spawnactor}.
In this appendix, we describe how this contract maps in principle to representative APIs across four runtimes: JavaScript (web stack), Godot, Unity, and Unreal Engine.
The examples below are illustrative mappings rather than engine-native implementations used in our experiments.

To make the discussion concrete, we use a running example throughout: constructing a boss fight state in which the boss is spawned in a designated arena, the player is positioned nearby with specific health and inventory, quest flags are advanced to the boss phase, and combat parameters are initialized.

\subsection{JavaScript / Web Stack}

Web-stack games (e.g., those built with Phaser, Three.js, or vanilla Canvas/DOM) run in a browser environment where game state is typically held in mutable JavaScript objects and can often be inspected or modified through the JavaScript runtime~\cite{chrome_devtools_runtime}.
Runtime state patching is comparatively straightforward when relevant state is exposed through the global scope, module exports, debug hooks, or lightweight instrumentation.

\begin{lstlisting}[language=JavaScript, basicstyle=\ttfamily\small, frame=single, caption={Runtime state patching in a web-stack game.}]
// Spawn boss in the arena
const boss = game.addEntity("BossEnemy", {
  position: { x: 500, y: 300 },
  archetype: "dragon",
  parent: game.scenes["bossArena"]
});

// Configure player and game state
game.player.position = { x: 480, y: 350 };
game.player.health = 100;
game.player.inventory.push("legendary_sword");
game.questManager.setFlag("boss_phase", true);
game.combat.bossPhase = 2;
game.combat.timer = 0;
\end{lstlisting}

Because JavaScript is dynamically typed and exposed runtime objects are mutable, LLM-generated web games often have less indirection between the evaluator and accessible game state.
This makes web-stack games the most straightforward target for our verification paradigm, which is why our current experiments focus on this runtime.

\subsection{Godot Engine}

Godot organizes game state as a tree of \texttt{Node} objects (the scene tree), where each node has typed properties accessible via GDScript or the Godot API~\cite{godot_nodes_scene_instances}.
Runtime state patching maps onto Godot's scene tree manipulation and property access.

\begin{lstlisting}[language=Python, basicstyle=\ttfamily\small, frame=single, caption={Runtime state patching in Godot (GDScript).}]
# Load arena and spawn boss
var arena = load("res://scenes/BossArena.tscn").instantiate()
get_tree().current_scene.add_child(arena)

var boss = load("res://prefabs/Dragon.tscn").instantiate()
boss.position = Vector2(500, 300)
arena.add_child(boss)

# Configure player and game state
var player = get_node("/root/Main/Player")
player.position = Vector2(480, 350)
player.get_node("Health").current_hp = 100
player.get_node("Inventory").add_item("legendary_sword")

GameManager.quest_flags["boss_phase"] = true
GameManager.combat_phase = 2
GameManager.combat_timer = 0.0
\end{lstlisting}

Godot's scene tree is accessible and mutable at runtime, and nodes can be instantiated, reparented, or removed programmatically~\cite{godot_nodes_scene_instances,godot_packedscene_class}.
Notably, community MCP (Model Context Protocol) servers have been developed for Godot that expose engine operations such as scene tree inspection, node creation, and script editing to external AI agents~\cite{godot_mcp_ee0pdt}.
However, these MCP integrations are still in an early stage of development.

\subsection{Unity Engine}

Unity organizes game state through \texttt{GameObject}s and \texttt{Component}s~\cite{unity_gameobject_getcomponent}.
Each GameObject lives in a scene hierarchy and carries components that define its behavior and data.
Runtime state patching maps onto Unity's programmatic scene management, object instantiation, and component property access~\cite{unity_object_instantiate,unity_gameobject_getcomponent}.

\begin{lstlisting}[language={[Sharp]C}, basicstyle=\ttfamily\small, frame=single, caption={Runtime state patching in Unity (C\#).}]
// Load arena scene, spawn boss
SceneManager.LoadScene("BossArena");

var bossPrefab = Resources.Load<GameObject>("Prefabs/Dragon");
var boss = Instantiate(bossPrefab,
    new Vector3(500, 0, 300), Quaternion.identity);
boss.transform.SetParent(
    GameObject.Find("ArenaRoot").transform);

// Configure player and game state
var player = GameObject.FindWithTag("Player");
player.transform.position = new Vector3(480, 0, 350);
player.GetComponent<Health>().currentHP = 100;
player.GetComponent<Inventory>().AddItem("legendary_sword");

GameManager.Instance.questFlags["boss_phase"] = true;
GameManager.Instance.combatPhase = 2;
GameManager.Instance.combatTimer = 0f;
\end{lstlisting}

Unity's Prefab system provides a natural mapping for controlling what entities exist: prefabs serve as archetypes that can be instantiated at arbitrary positions in the scene hierarchy.
Controlling what values entities hold is supported through direct component field access or, for more generic injection, through C\# reflection.
Overall, Unity's runtime provides primitives for runtime state patching through scene management, object instantiation, and component access.

\subsection{Unreal Engine 5}

Unreal Engine organizes game state through \texttt{Actor}s placed in a \texttt{UWorld}, with behavior defined by \texttt{UActorComponent}s~\cite{unreal_actors}.
State patching maps onto Unreal's actor spawning, property system, and C++ or Blueprint scripting~\cite{unreal_spawnactor}.

\begin{lstlisting}[language=C++, basicstyle=\ttfamily\small, frame=single, caption={Runtime state patching in Unreal Engine 5 (C++).}]
// Spawn boss in the arena
FActorSpawnParameters SpawnParams;
SpawnParams.Owner = ArenaRoot;
ABossEnemy* Boss = GetWorld()->SpawnActor<ABossEnemy>(
    BossClass,
    FVector(500.f, 300.f, 0.f),
    FRotator::ZeroRotator,
    SpawnParams);

// Configure player and game state
APlayerCharacter* Player = Cast<APlayerCharacter>(
    UGameplayStatics::GetPlayerCharacter(this, 0));
Player->SetActorLocation(FVector(480.f, 350.f, 0.f));
Player->HealthComponent->CurrentHP = 100.f;
Player->InventoryComponent->AddItem(
    TEXT("legendary_sword"));

AMyGameState* GS = GetWorld()->GetGameState<AMyGameState>();
GS->QuestFlags.Add(TEXT("boss_phase"), true);
GS->CombatPhase = 2;
GS->CombatTimer = 0.f;
\end{lstlisting}

Unreal's property system (\texttt{UPROPERTY} macro) exposes component fields to both the editor and runtime reflection, enabling generic state patching without hardcoded field access~\cite{unreal_object_handling}.
Like Unity, Unreal's runtime supports programmatic control over the actor hierarchy and component state, providing primitives for runtime state patching.

\subsection{Summary}

These runtimes provide primitives that can support the two fundamental capabilities required by our injection contract: controlling what entities exist in a running scene and controlling what values they hold~\cite{chrome_devtools_runtime,godot_nodes_scene_instances,unity_object_instantiate,unreal_spawnactor}.
The key difference lies in accessibility: web-stack games expose state with less indirection, while game engines require going through their respective object models (scene tree, GameObject/Component, Actor/Component).
Since Godot, Unity, and Unreal Engine do not yet support mature LLM-based game generation (community MCP servers for these engines have begun to appear but remain early-stage~\cite{godot_mcp_ee0pdt,unity_mcp_coplaydev,unreal_mcp_chir24}), they are outside the scope of our current experiments.
However, as demonstrated above, all three expose runtime state patching primitives, meaning our verification paradigm could in principle be extended to these engines with engine-specific adapters once their generation pipelines mature~\cite{godot_nodes_scene_instances,unity_object_instantiate,unreal_spawnactor}.

\section{Agent-as-a-Verifier Direct Baseline Prompt}
\label{app:AaaV_prompt}

We provide the user prompt used for the Agent-as-a-Verifier (AaaV) baseline below.
The agent is given an anonymized game identifier, the corresponding specification path, and a target run identifier.
The system prompt is empty, so the baseline executes purely from the user prompt without any additional skill scaffolding.
The prompt template is:

\begin{lstlisting}[basicstyle=\ttfamily\footnotesize,breaklines=true,columns=fullflexible]
Evaluate whether the game implementation is correct.

Hard requirements:
- Your current working directory is the generated implementation directory for anonymized game identifier `{game_id}`. The primary specification file is `{specification_path}`. If it is missing, say so explicitly in the result.
- You must choose your own testing path, but you must write a structured record of what you actually tested.
- Write `baseline_eval_result.json` in the current working directory.
- All output text must be in English, including `checks`, `findings`, `limitations`, and `summary`.
- Keep the evaluation bounded and pragmatic. Do not keep exploring once you already have decisive evidence for the final verdict.
- Required workflow:
  1. Read the full specification end-to-end. Enumerate every concrete specification element, where an element may be a sentence-level or clause-level requirement covering inputs, physics/timing, win conditions, lose conditions, state transitions, game rules, HUD/UI requirements, boundary conditions, or other categories.
  2. Run `npm install` if needed, then run one build command.
  3. Launch one temporary local server with the lifecycle helper.
  4. Write `baseline_eval_result.json` and stop.
- Do not perform open-ended exploration loops, repeated browser sessions, or speculative deep dives after every element already has a verdict.
- Do not invent elements that are not in the spec.
- The JSON must include at least:
  - `game_id`
  - `run_id`
  - `final_verdict` (`pass` | `fail`)
  - `confidence` (0-100)
  - `checks_performed` (string list describing the checks you actually performed)
  - `files_inspected` (string list)
  - `commands_ran` (string list)
  - `findings` (object list; each item must include at least `severity`, `title`, `evidence`)
  - `limitations` (string list explaining what you could not verify and why)
  - `summary`

- If some logic could not be verified because the entry path was blocked, the interaction path was unreachable, or evidence was insufficient, say that in `limitations`. Do not pretend the check was completed.
- If you ran build steps, launched the page, used a Playwright smoke test, or performed static inspection, include them in both `checks_performed` and `commands_ran`.
- If you need a local server, do not run `npm run dev` or `npm run preview` in the foreground.
- Instead, you must use this exact lifecycle helper pattern:
  - Start:
    ```bash
    eval "$("<LIFECYCLE_HELPER>" start --identifier "{game_id}" --timeout-seconds 30)"
    ```
  - Use `$GAME_URL` for browser checks.
  - Stop before exit:
    ```bash
    "<LIFECYCLE_HELPER>" stop --state-file "$SERVER_STATE_FILE"
    ```
- Do not finish while a dev server is still running.
- If you did not perform a check, do not fabricate it.
- Finish by printing one short terminal line that mentions `baseline_eval_result.json`.

Parameters:
- `game_id={game_id}`
- `specification_path={specification_path}`
- `run_id={run_id}`
\end{lstlisting}

\section{Review-and-refine Loop}
\label{app:refine}

Not every initially generated verification unit is immediately executable.
After generation, each unit undergoes a review step in which the agent checks whether the unit is well-formed: whether the state $s_k$ is constructible, whether the instruction $i_k$ is unambiguous, and whether the expected outcome $y_k$ admits a clear pass or fail determination.
If any of these conditions are not met, the unit is refined iteratively until it yields a definitive verdict.
This review-and-refine loop ensures that every verification unit entering the parallel execution stage produces a conclusive result, avoiding ambiguous outcomes that would undermine the reliability of the aggregate score.

\section{Data Synthesis Pipeline}
\label{sec:data}

To construct a benchmark for \name{}, we build a dataset of game generation tasks, each consisting of an anonymized game identifier and a structured natural-language game specification.
The pipeline proceeds in three stages.

\paragraph*{Stage 1: Sourcing.}
We source candidate games from popular digital game marketplaces, specifically the Apple App Store~\cite{apple_app_store_games} and Steam~\cite{steam_store}.
We collect a broad initial pool of games across seven target genres: Action, Adventure, Casual, Puzzle, Simulation, Strategy, and Board.
We choose these genres because they are common marketplace-level game categories in the Apple App Store Games catalog and Steam's genre/tag taxonomy, with Steam representing board games through Board Game/Tabletop tags~\cite{apple_app_store_games,steamworks_tags}.
Games are initially filtered by popularity and quality, retaining only titles with substantial player engagement and high average ratings.
We then apply rubrics-based LLM scoring.

\textit{Rubrics-based suitability scoring.}
Each candidate that passes the hard filters is scored by an LLM judge (GPT-5.4) across four rubrics on a 1--5 scale:
\begin{denseitemize}
    \item \textbf{Textual Reconstructability}: whether the game's core mechanics, rules, and progression can be fully specified in natural language such that the specification alone is sufficient for an LLM to produce a functionally correct implementation.
    \item \textbf{Rule Determinism}: whether the game's rules are well-defined and unambiguous.
    \item \textbf{Mechanic Diversity}: whether the game exercises multiple distinct logic mechanisms (e.g., collision, scoring, state transitions, AI behavior, phase progression), providing sufficient variety for comprehensive evaluation.
    \item \textbf{Bounded Complexity}: whether the game's scope falls within the generation capacity of current LLM-based coding agents, excluding titles whose scale (e.g., AAA-level content volume or engine complexity) makes faithful generation infeasible.
\end{denseitemize}
We retain games scoring at least 3 on every rubric and rank candidates by their aggregate score.
The final curated set contains 100 games spanning all seven genres.

\paragraph*{Stage 2: Specification synthesis.}
For each selected game, GPT-5.4 with deep search aggregates publicly available information and synthesizes a structured game specification.
Each specification defines the core objective, controls, entities, rules, scoring, progression, failure conditions, and major mechanics.
The specification abstracts away platform-specific assets and UI details while preserving the core gameplay logic, so that the specification alone is sufficient for an LLM to generate a functionally correct implementation.

Because the selected games are popular titles likely present in LLM training corpora, exposing the original game name during generation or evaluation could allow the model to rely on memorized knowledge rather than faithfully following the provided specification.
To mitigate this data contamination risk, we anonymize all game identities: each game is assigned a unique identifier (e.g., \texttt{Game-042}), and all references to the original game name are removed from the specification.
Both generation and evaluation operate exclusively on the anonymized identifier and specification.

\paragraph*{Stage 3: Human refinement.}
Human annotators refine each specification for accuracy, completeness, consistency, and evaluability.
This step removes ambiguity, normalizes terminology, and makes latent game logic explicit for both generation and evaluation.

\paragraph*{Final dataset: \textsc{VeriGame}.}
The final dataset contains 100 examples spanning seven genres.
Each example contains an anonymized game identifier and a game specification.